# State Classification with CNN

Astha Sharma

University of South Florida

*Email: asthasharma017@gmail.com*

## I. ABSTRACT

There is a plenty of research going on in field of object recognition, but object state recognition has not been addressed as much. There are many important applications which can utilize object state recognition, such as, in robotics, to decide for how to grab an object. A convolution neural network was designed to classify an image to one of its states. The approach used for training is transfer learning with Inception v3 module of GoogLeNet used as the pre-trained model. The model was trained on images of 18 cooking objects and tested on another set of cooking objects. The model was able to classify those images with 76% accuracy.

## II. INTRODUCTION

Image classification is the basis of various applications. From facial expression recognition to identification of position of stars in galaxy, image classification is used everywhere. In past few years, there has been an evolution in the methods of image classification. Some very effective and popular methods have been developed such as, Support Vector Machine, Artificial Immune System, Deep Learning, etc. [1]. Deep learning has been proved as one of the most effective technique for machine learning. In deep learning, multiple layers of neurons are built to train a network. All the pretrained networks use dozens of layers which builds thousands of parameters in the network. Complexity, overfitting and requirement of large dataset are downsides of such very deep networks.

Object state classification is very useful in various applications in fields like robotics, medical, astronomy, gaming and many more. We studied some previous work done in robotics which utilizes object state recognition. Recognizing an object's state is important in robotics. Objects in different states requires different grasping which will require distinct manipulations in grasping action of a robot. A new structure of knowledge representation is functional object-oriented network (FOON) [2]. FOON is a graphical model learned with objects in different states and motion of their manipulation operations. FOON consist of various subgraphs demonstration different object states and their manipulation movement. This can be used by robots to learn object state recognition.

Choosing an appropriate grasp and manipulation of motion have been active research areas in robotics. Object state recognition plays an important role in determination of grasp. One way of handling this problem is using shape matching algorithms [3].

Another approach is object–object-interaction affordance knowledge to perform object classification and action recognition. "This method can connect and model the motion and features of an object in the same frame" [4]. The training is done with labeled video sequences and represented as a Bayesian Network. Bayesian Network includes objects, human action, and object reaction as parameters [4].

This report concentrates image state classification using deep convolution neural network and a pretrained network for transfer learning. The paper is based upon the experiments performed to classify states of multiple categories of different objects. We will discuss techniques which are used in our experiment to avoid overfitting as well as to train a network efficiently using limited dataset.

The experiments were performed on the set of 5177 training images and 861 testing images with 7 categories of classes. The task was to classify the state of a given object. We designed a convolution neural network which classifies an image to one of its states. Inception v3 module of GoogLeNet was used for transfer learning. Data augmentation and rescaling of input is done as preprocessing. In the convolution neural network, multiple layers of convolution, normalization, pooling, activation, flattening and dropout are used. The model was optimized twice using RMSprop and Stochastic Gradient Descent (SGD) optimizers.

## III. DATA COLLECTION AND PREPROCESSING

Deep learning uses neural networks with hundreds of hidden layers and it requires large amounts of training data. Building a good neural network model always requires careful consideration of the architecture of the network as well as the input data format. Data collection and preprocessing are very crucial parts of deep learning related experiments.

For doing data collection, data annotation was done by students of Deep Learning class of University of South Florida. Every student was given 2 batches of videos and images. In the first batch each student was assigned 10-20 short videos (1-minute video with 0.5-1 fps) of only a single object (for example: onion). The student had to draw the bounding box for each object of the type (onion) in each frame of the video and annotate the bounding box with the state of that object (for example: sliced). In the second batch, each student was assigned a set of images of the same object (onion). The student had to label each image with the state label (sliced). After annotations were done, each student was given a random set of videos and images from other students to check for annotation errors.

Pre-processing of data is a technique that involves transforming raw data into a managed and consistent format. Data can be incomplete, inconsistent, or lack certain behaviors [5]. Data preprocessing helps to resolve such issues. This technique prepares raw images data to be processed further as the input to the neural network [6].

The data used in this project is the images of 18 cooking objects (tomato, onion, chees, bread, egg, etc.) with 7 different states (whole, diced, sliced, julienne, grated, paste, juiced). There were approximately 5177 images provided as training dataset. First step of preprocessing is to partition data into training and validation datasets. After partitioning, multiple pre-processing techniques like resizing, image scaling, data augmentation and normalization were performed on the images in dataset.

Data Partitioning: Data must be partitioned into training and validation datasets. A good practice for partitioning data is to shuffle data before partitioning so that each training and validation sets have equal amount of data in all categories. For this project, data partitioning is done with 4:1 ratio. 80% of data is kept for training and remaining 20% for validation. Out of 5177 images, approximately 3,882 images are used for training and 1294 images are used for validation purpose.
After partitioning, input data is fed to ImageDataGenerator, which takes care of data normalization and augmentation. It generates batches of normalized and augmented data. Class modes used can be ‘categorical’, ‘binary’, ‘sparse’, ‘input’ or None depending upon classes of data. There are 7 classes of data in this project, so, the class_mode used is 'categorical'.

Input Image Resizing: When input image size is large, it increases the amount of noise and variance. That could mean more convolution and pooling layers because the network must deal with it to process input. The network may also need more training examples, and each training example will be large. The large size of image increases the computation and make the network slow and complex. On the other hand, too small image sizes may cause losing some important attributes to be used to classification (Figure 1) [7].

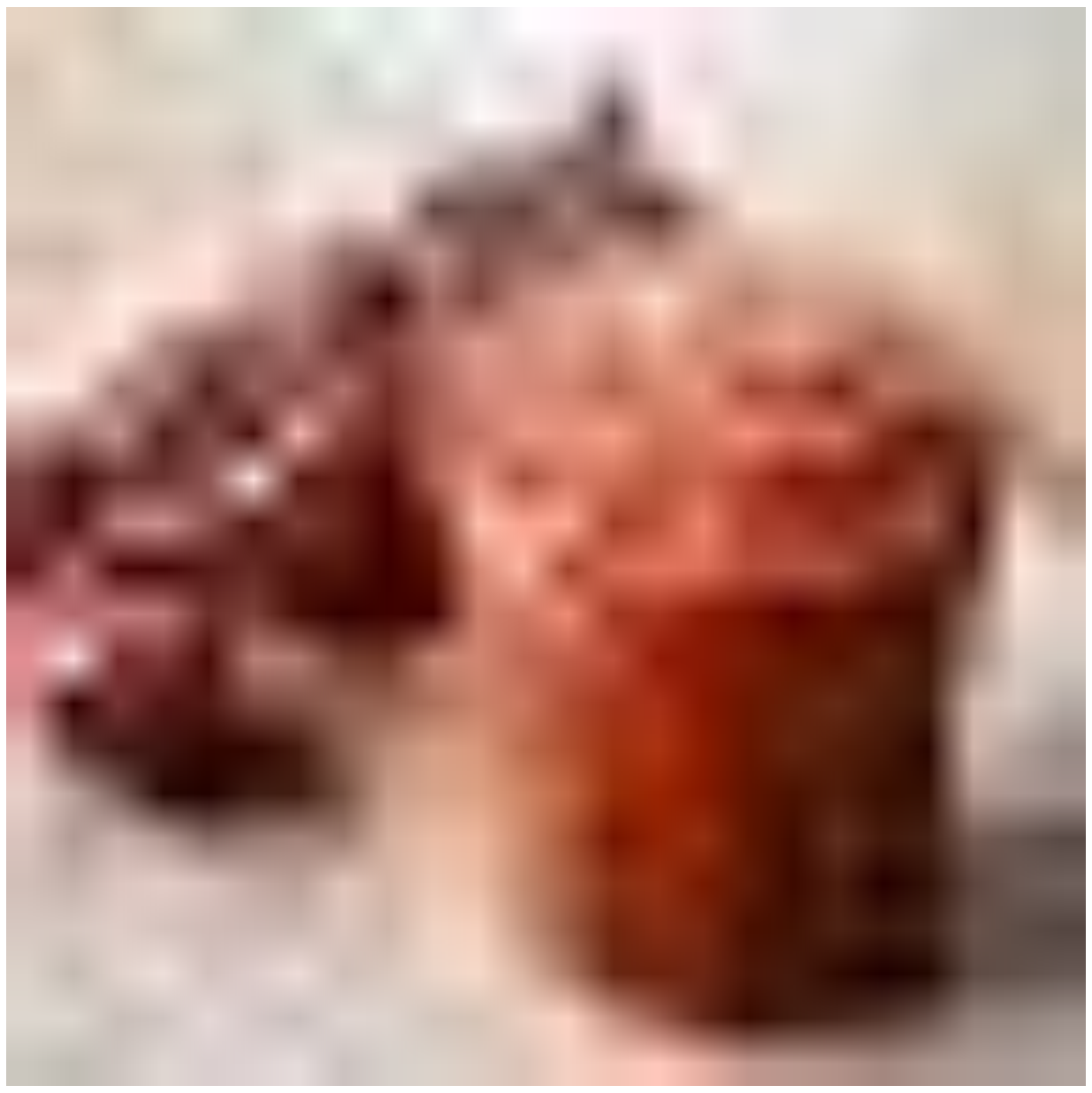

Figure 1: Input size too low for classification attributes to be detected correctly.

We trained the model on different image sizes and for different number of epochs. It was observed that if the images size is larger, the model needs more number of epochs to reach to a good accuracy. After doing experiments by training the model keeping different image sizes, image size 363*363 was finalized for the input images as it was giving the best accuracy with reasonable number of epochs.

Image Scaling: Rescaling an image resizes it by a given scaling factor. Image rescaling is an important technique in data preprocessing and can impact the accuracy highly. We can up-scale and down-scale the images and there are few library functions to do this. Usually downscaling is done for preparing the image to be fed in neural network. In this experiment, image downscaling is used with a scaling factor of 1/255 [8].

Normalizing image inputs: Normalization ensures that each input parameter has a similar data distribution. Hence it makes the convergence of the network faster while training [9]. Normalization of data is done by subtracting mean from each pixel and dividing that result by the standard deviation. In this experiment, images were normalized in the range [0,255].

Data augmentation: If we have low amount of data in our dataset, data augmentation can be used as an effective pre-processing technique. It involves augmenting the existing data-set by scaling, rotations (Figure 2), zoom, fill mode etc. Doing data augmentation, we can get a wide verity of data. If a neural network is trained on small dataset, it may be overfitted or may not perform well over test data. Data Augmentation gives advantage by increasing the size of dataset. This can help preventing over-fitting.

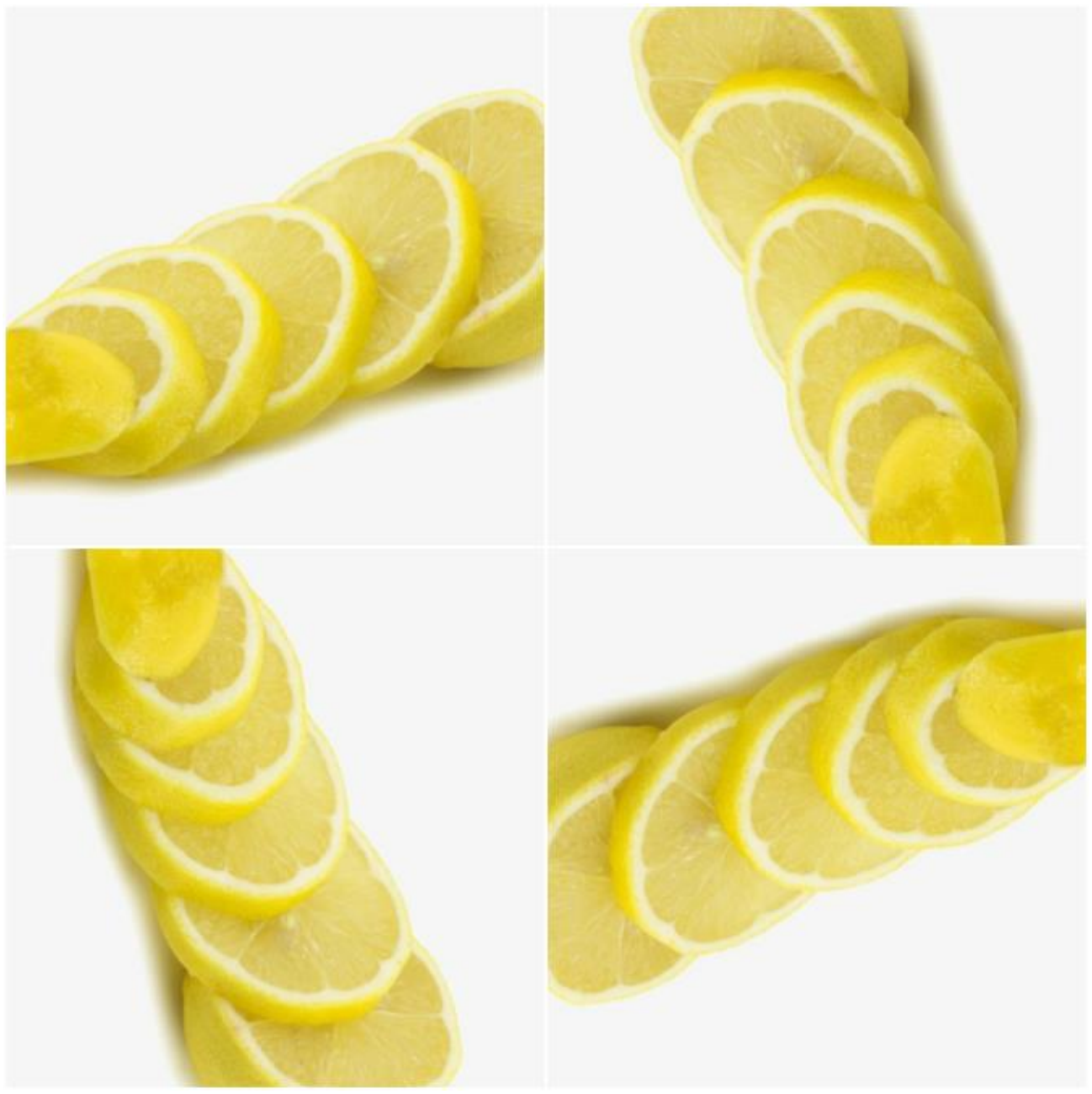

Figure 2: Data Augmentation using horizontal_flip, vertical_flip and rotation_range.

There are 9 types of offline data augmentation done on the input data for this project. The parameters used for augmentation are rotation range, width shift range, height shift range, rescale, shear range, zoom range, horizontal flip, vertical flip and fill mode. Total size of data set after augmentation is approximately 28,000 images.

## IV. Methodology

### A. Base Model

The method used for training the network is transfer learning. In real-world applications, sometimes there can be situations where there is no sufficient data to train a network for effective results. In such cases we can use transfer learning. The training used for pretrained network may be sufficient only for one domain, and the data we have may be in a different feature space. In these situations, knowledge transfer, would greatly improve the learning performance of neural network by avoiding data-labeling which is very expensive and takes lots of efforts. Transfer learning is a new and effective learning framework to address this problem [10].

Pretrained model used in this experiment is Inception v3. Inception v3 is a deep convolutional neural network architecture based on GoogLeNet and developed by Google (Figure 3) [12]. This architecture achieves the new accuracy for classification and detection in the ImageNet Large-Scale Visual Recognition Challenge 2014 (ILSVRC14) [11]. Inception performs the utilization of the computational resources in an improved way. This network has improved depth and width while keeping lower computation budget [11][12]. “The architectural decisions were based on the Hebbian principle and the intuition of multi-scale processing” [11]. The model is 22 layers deep and known to give higher accuracy than any other pretrained models [11].

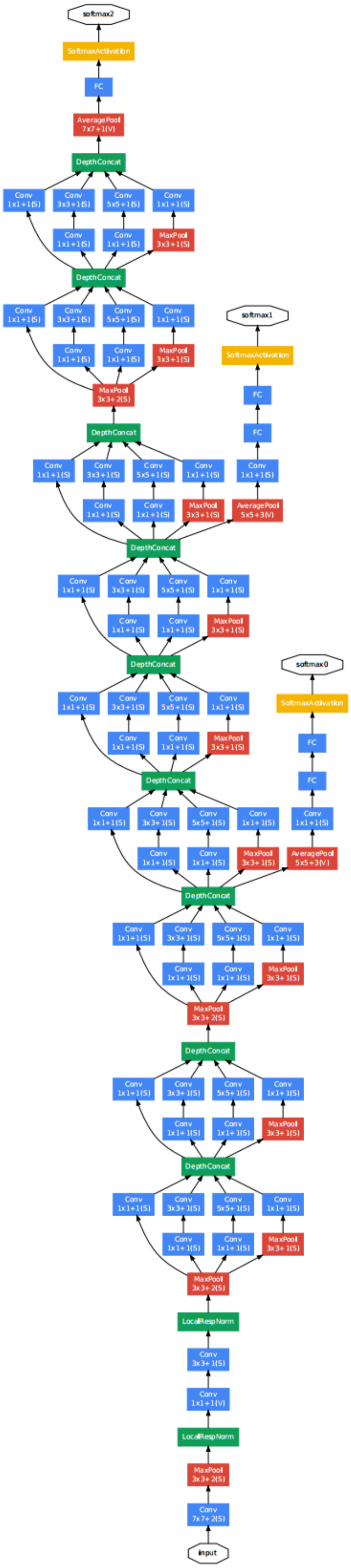

Figure 3: GoogleNet, base model architecture [12]

### B. Proposed Model

The model created to be trained on the new dataset in this project has some layers added on top of the pretrained model. In this experiment 16 new layers have been added to the network (Figure 4).

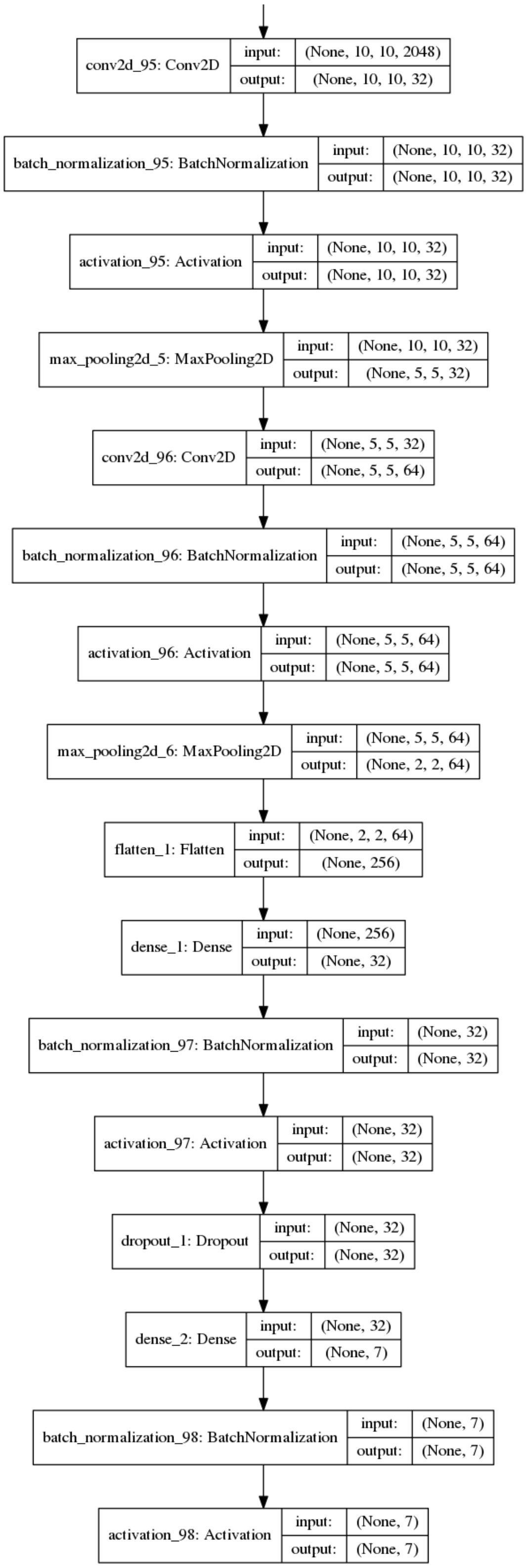

Figure 4: Architecture of top layers of the proposed model

First layer of the network is a convolution layer. A convolution layer generates output in form of array of tensors as a convolution kernel convolved with the input of this layer. Input shape and dimensions must be determined if this layer is used as the first layer of network [13] [14].

The next layer is Batch Normalization layer. Batch Normalization provides a way to shift inputs closer to zero or mean of all values in the input array. This can be used at the beginning as well as in middle of the layers. This technique helps in faster learning and getting higher accuracy. In the deep convolution neural network, as the data flows further, the weights and parameters adjust their values. This flow can make the intermediate data too big for computation or too small to give correct prediction. This problem can be avoided by normalizing data in each mini batch [15].

Next layer used in the network is Activation layer with Rectified Linear Unit (ReLU) as the activation function. ReLU has become very popular in recent years and it is proved to have 6 times improvement in convergence from Tanh function [16]. Mathematical formula of ReLU is:

$$R(x) = max(0,x)$$
$$i.e \text{ if } x < 0 \text{ , } R(x) = 0$$
$$\text{and if } x >= 0 \text{ , } R(x) = x$$

ReLU used as an activation function have been highly successful for computer vision tasks and gives better speed and performance than standard sigmoid function [16].

Another layer is max pooling. The objective of doing max pooling is down-sampling of data, which avoids overfitting by reducing the dimensionality of data. Another benefit of max pooling is that it reduces computational cost by decreasing number of parameters [17]. In this experiment, MaxPooling2D function of Keras library is used twice for max pooling. It is reducing significantly dimensionality of data once from 10*10 to 5*5 and then from 5*5 to 2*2 (Figure 4). Two sets of convolutions, batch normalization, activation and max pooling layers are used in the network for getting higher accuracy.

The next layer is Flatten. Flattening converts the 2-dimensional output of previous layer into a single long continuous linear vector. This is required before the flow goes to fully connected or dense layer. A dense layer is just an artificial neural network (ANN) classifier and requires individual features. Flattening is needed to convert the input array into a feature vector [18].

Next layer of the network is a fully connected or dense layer. A dense layer is a fully connected neural network layer in which each input node is connected to each output node. It takes a flattened vector as input and gives n-dimension tensor as output. Using dense layer has several advantages. “It solves the vanishing-gradient problem, strengthen feature propagation, encourage feature reuse, and substantially reduce the number of parameters” [19]. There are two fully connected layers used in the model.

The next layer is dropout, which is used for regularization. Regularization helps the model to generalize better by reducing the risk of overfitting. There is a risk of overfitting if the size of data set is too small as compared to the number of parameters needed to be learned. A dropout layer randomly removes some nodes and their connections in the network. We can use dropout with hidden or input layer. By using dropout, the nodes in the network become more insensitive to the weights of the other nodes. This makes the model is robust [16] [20].

Last layer in the network is the activation layer with softmax activation function. Softmax function converges the output between 0 and 1. It is usually better to use softmax in the last layer of network for classification problems. [21]. It is used in categorial probabilistic distribution. Mathematical formula of softmax function is:

$$\sigma(z)_j = \frac{e^{z_j}}{\sum_{k=1}^{K} e^{z_k}}$$

Summary of the neural network model is shown below (Figure 5).

| Layer (type) | Output Shape | Param # | Connected to |
|---|---|---|---|
| conv2d_95 (Conv2D) | (None, 10, 10, 32) | 589856 | mixed10[0][0] |
| batch_normalization_95 | (None, 10, 10, 32) | 128 | conv2d_95[0][0] |
| activation_95 (Activation) | (None, 10, 10, 32) | 0 | batch_normalization_95[0][0] |
| max_pooling2d_5 (MaxPooling2D) | (None, 5, 5, 32) | 0 | activation_95[0][0] |
| conv2d_96 (Conv2D) | (None, 5, 5, 64) | 18496 | max_pooling2d_5[0][0] |
| batch_normalization_96 | (None, 5, 5, 64) | 256 | conv2d_96[0][0] |
| activation_96 (Activation) | (None, 5, 5, 64) | 0 | batch_normalization_96[0][0] |
| max_pooling2d_6 (MaxPooling2D) | (None, 2, 2, 64) | 0 | activation_96[0][0] |
| flatten_1 (Flatten) | (None, 256) | 0 | max_pooling2d_6[0][0] |
| dense_1 (Dense) | (None, 32) | 8224 | flatten_1[0][0] |
| batch_normalization_97 | (None, 32) | 128 | dense_1[0][0] |
| activation_97 (Activation) | (None, 32) | 0 | batch_normalization_97[0][0] |
| dropout_1 (Dropout) | (None, 32) | 0 | activation_97[0][0] |
| dense_2 (Dense) | (None, 7) | 231 | dropout_1[0][0] |
| batch_normalization_98 | (None, 7) | 28 | dense_2[0][0] |
| activation_98 (Activation) | (None, 7) | 0 | batch_normalization_98[0][0] |

Total params: 22,420,131
Trainable params: 617,077
Non-trainable params: 21,803,054

Figure 5: Summary of the proposed model

### C. *Training*

Training of the model is done in two steps. At first, only the newly added layers are trained on given dataset keeping the base model layers non-trainable. After training new layers, fine tuning is done by training top four layers of base model and keeping rest of the bottom layers non-trainable. Experiments were done on three optimizers: RMSprop, SGD and Adam. Best results were obtained using combination of RMSprop and SGD optimizers. Firstly, the model is compiled with the optimizer RMSprop training only new layers and keeping the base layers non-trainable. RMSprop works by dividing the learning rate by an exponentially decaying average of squared gradients [23]. Default value of learning rate for RMSprop is 0.001 [22]. The objective function used to compile the model is 'categorical_crossentropy', which is commonly used when number of classes on the dataset is more than two. Experiments were done with keeping number of epochs as 20, 50 and 100 while keeping the batch size as 32. The experiment had to be done on comparatively smaller dataset and large number of epoch. This might cause overfitting. Hence, batch size (32) was chosen smaller to avoid overfitting. The best model (in terms of validation loss and accuracy) was saved during the training so that it can be used for performing training in the second step.

In the second step of training, fine-tuning was done by training the top 4 layers of base model with the given dataset, while freezing the remaining bottom layers. The best model obtained in the first step of training was used in this process. Experiments were done with training top 2 and top 4 layers of base model in this step and training top 4 layers give best results. Optimizer used in this step was Stochastic Gradient Descent (SGD). Adding top 4 layers of Inception caused increased number of learning parameters, increasing chances of overfitting. SGD is used here because it is comparatively slower than other optimizers [22] and a slow convergence was needed to avoid overfitting. Default values of learning rate (0.0001), decay=1e-6 and momentum=0.9 were used with SGD. The training was done for further 100 epochs while saving all the models generated every epoch. The best model in terms of validation loss and accuracy was chosen which also had the lowest timestamp. Choosing the best model with lowest timestamp avoided overfitting. This approach of choosing the best model helps because there are chances that the model stops progressing with time, producing an over-fitted model. It was observed during training that model stopped progressing after 31 epochs. Hence the model generated at 31st epoch was chosen as the best model, manually stopping the training as compared to using early stopping where only the last generated model could be saved, which could have risk of overfitting. The total number of parameters trained during the process were 22,420,131.

### D. *Input and Output*

Input provided to the network is a 4-dimenssional array of tensors representing images in the dataset with the dimensions representing channel, height, width and number of images sequentially. The output produced by softmax function in the last layer of network is an array of tensors representing result as the probability of all the classes in one-hot notation. The class index with highest probability is the predicted class.

## V. EVALUATION AND RESULT

Various experiments were done during training to produce the best model. Few of the experiments are 1). training with one set of convolutional, batch normalization, activation and max pooling layers, 2). two sets of convolutional, batch normalization, activation and max pooling layers, 3). without fine tuning GoogleNet, and 4). fine tuning GoogleNet, 5). fine tuning GoogleNet with the best model produced by the training in step 1 (training with freezing all layers of base model). Results obtained with these experiments are shown below (Table 1).

| Training Options | Training Accuracy | Training Loss | Validation Accuracy | Validation Loss | #epochs |
|---|---|---|---|---|---|
| Training with one set of convolutional, batch normalization, activation and max pooling layers | 63% | 1.89 | 65% | 1.84 | 50 |
| Training with two sets of convolutional, batch normalization, activation and max pooling layers without fine tuning GoogleNet | 71% | 1.0 | 70% | 1.0 | 100 |
| Training with two sets of convolutional, batch normalization, activation and max pooling layers with fine tuning GoogleNet | 72% | 1.1 | 72% | 1.23 | 100 + 50(Fine tuning) |
| Training with two sets of convolutional, batch normalization, activation and max pooling layers with fine tuning GoogleNet with the best model produced by the training in step 1 | 75% | 0.86 | 74% | 0.87 | 100 + 31(Fine tuning) |

Table 1: Results generated with various experiments done during training.

From these experiments, it can be determined that adding more layers helps improving the results. Best results were obtained when the set of layers was fine-tuned with the pretrained network, GoogleNet in this case. Fine tuning is done with an already trained model, which was the best model chosen from previous training, further improves the results. Graphs to show training and validation accuracies and losses are shown below (Figure 6).

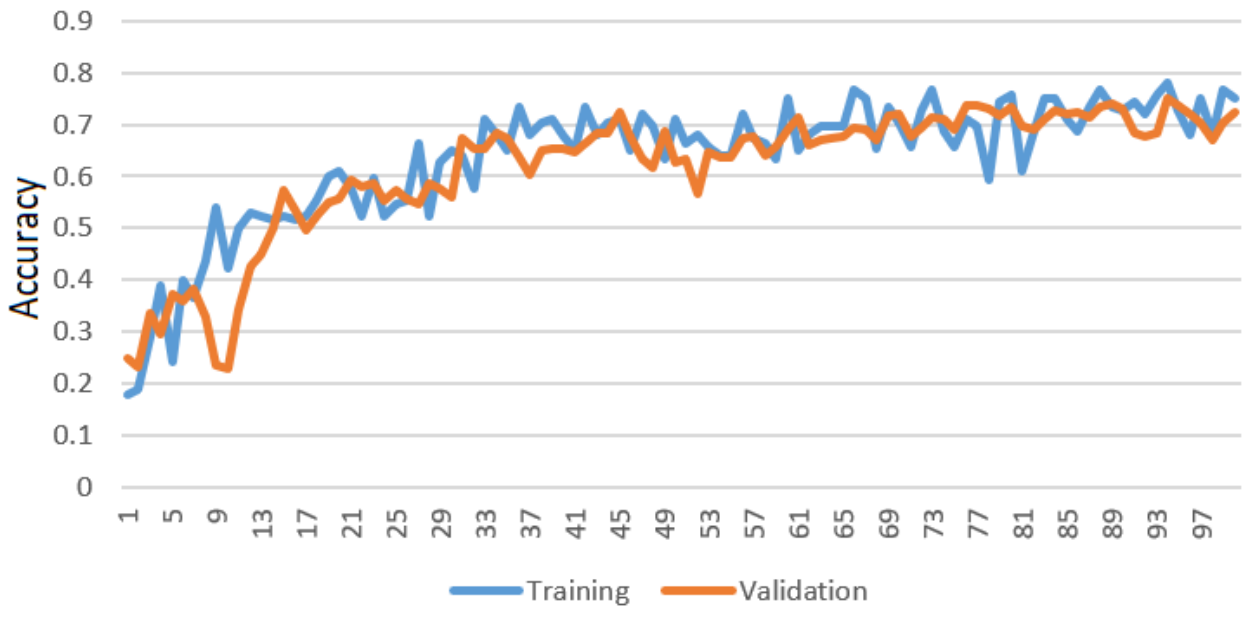

Epoch

Accuracies before fine tuning

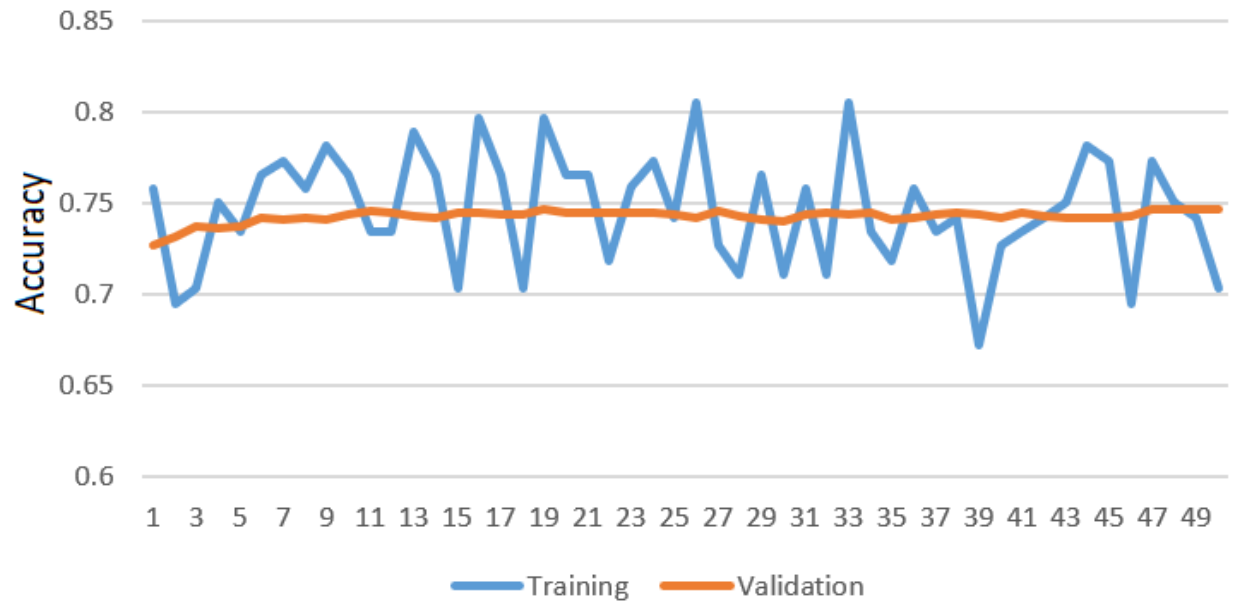

Epoch

Accuracies after fine tuning

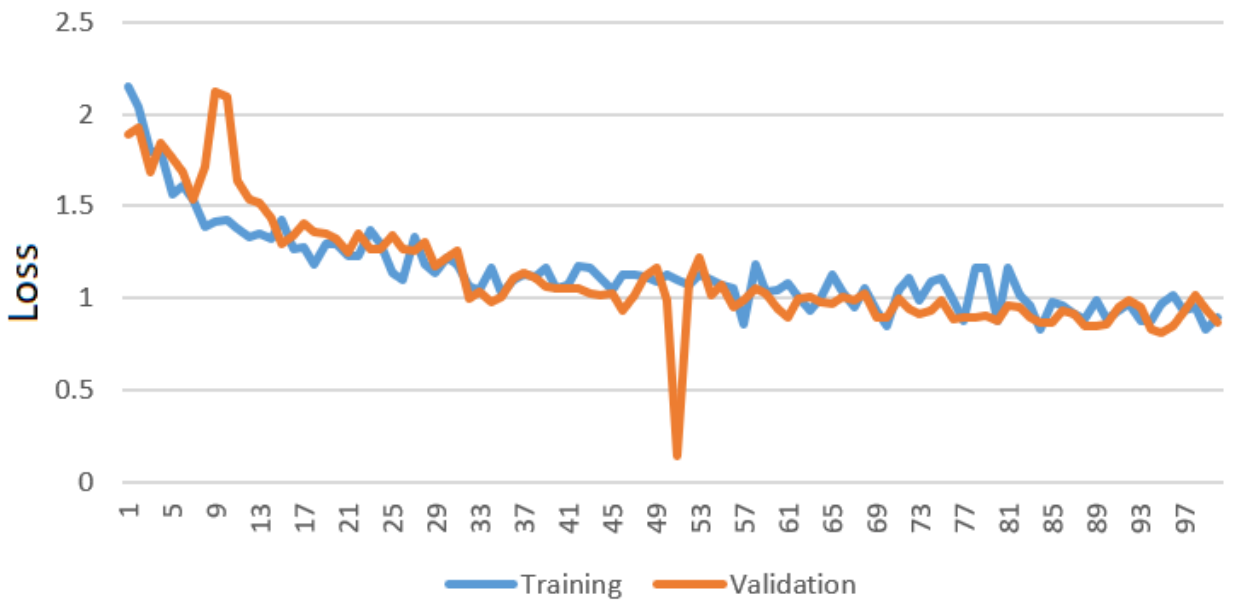

Epoch

Losses before fine tuning

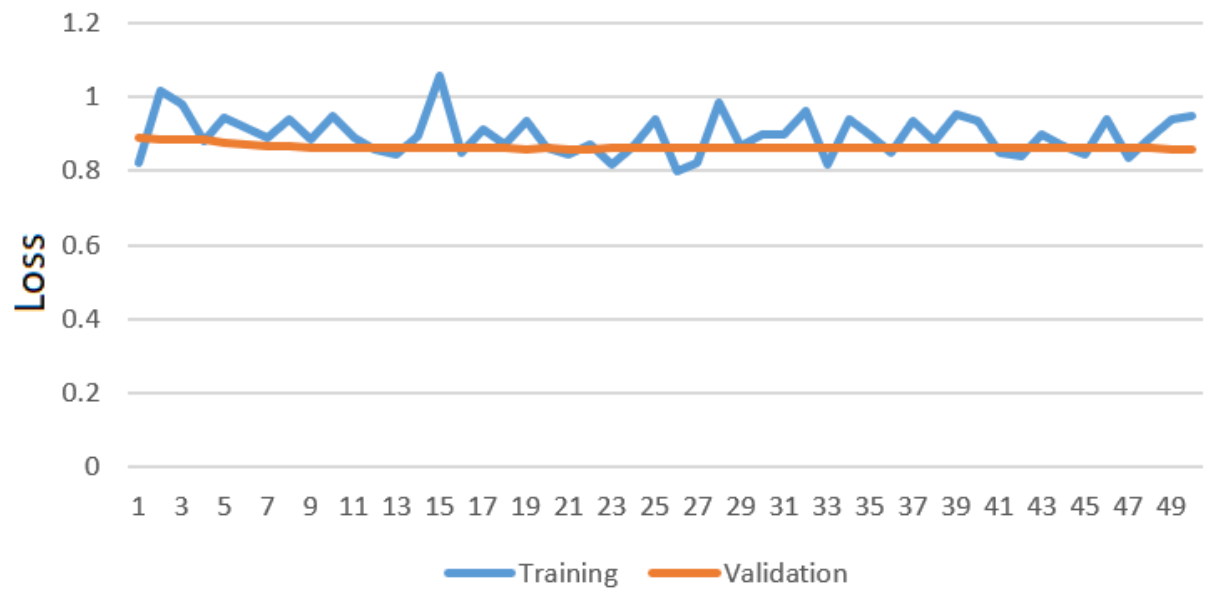

Epoch

Losses after fine tuning

Figure 6: Training and validation accuracies and losses before and after fine tuning.

From the graphs, it can be concluded that there is not much difference between training and validation results because both the training and validation graphs are converging almost simultaneously, in addition to it, both training and validation losses are consistently getting reduced. This means that the model is not over-fitted or underfitted. The test accuracy obtained with the model was 76%.

## VI. FUTURE WORK

There are few shortcomings in the experiments which impacted the results. Upon observing the data, it can be concluded that some of the data is noisy, which can be mislabeled easily. If these images are included in training database, it can result in wrong labeling while training and hence generating less accurate model. Model can be improved by eliminating such data from the dataset. Below are some examples of ambiguous data (Figure 8).

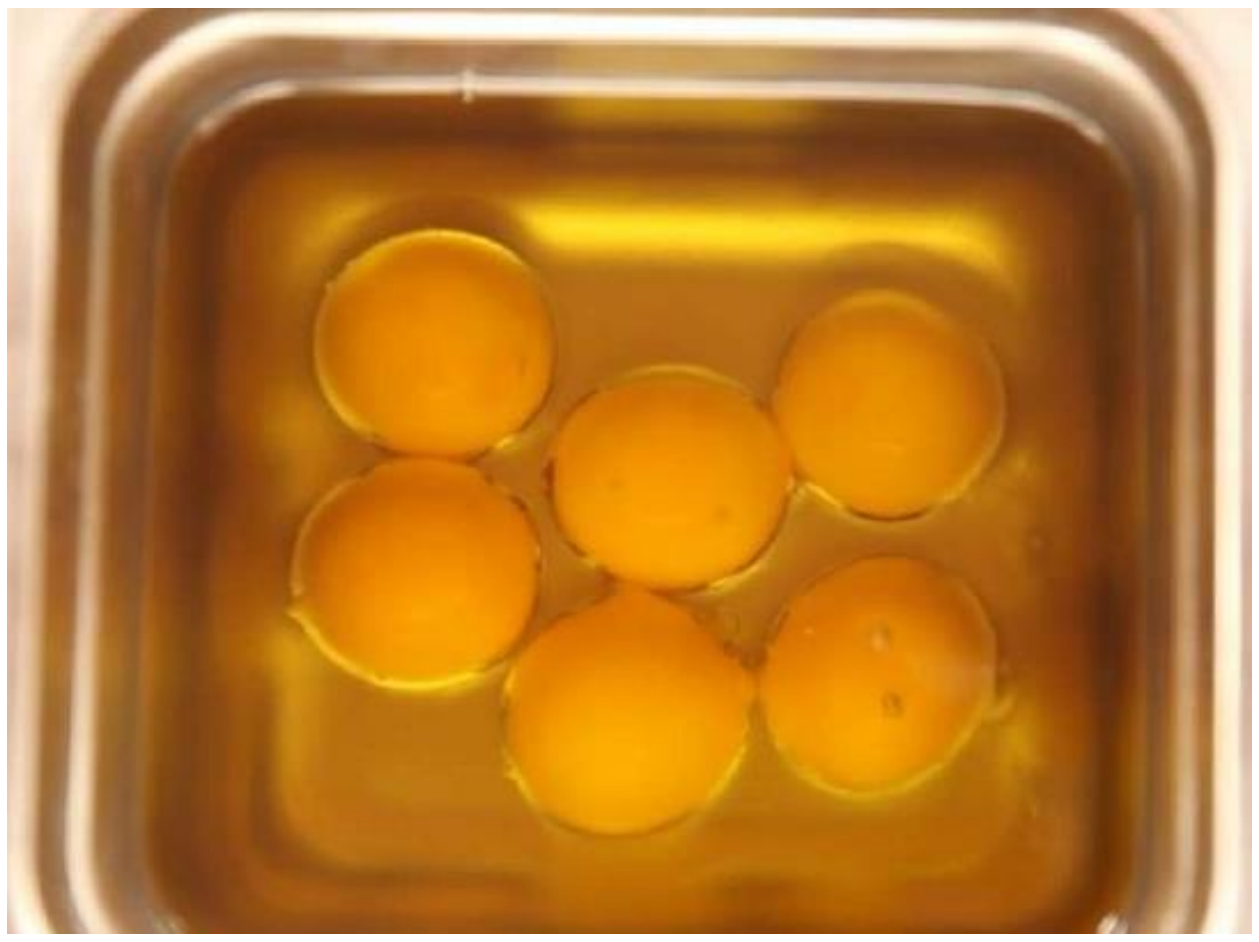

Ambiguous image: ambiguity between Juice, whole or slice states

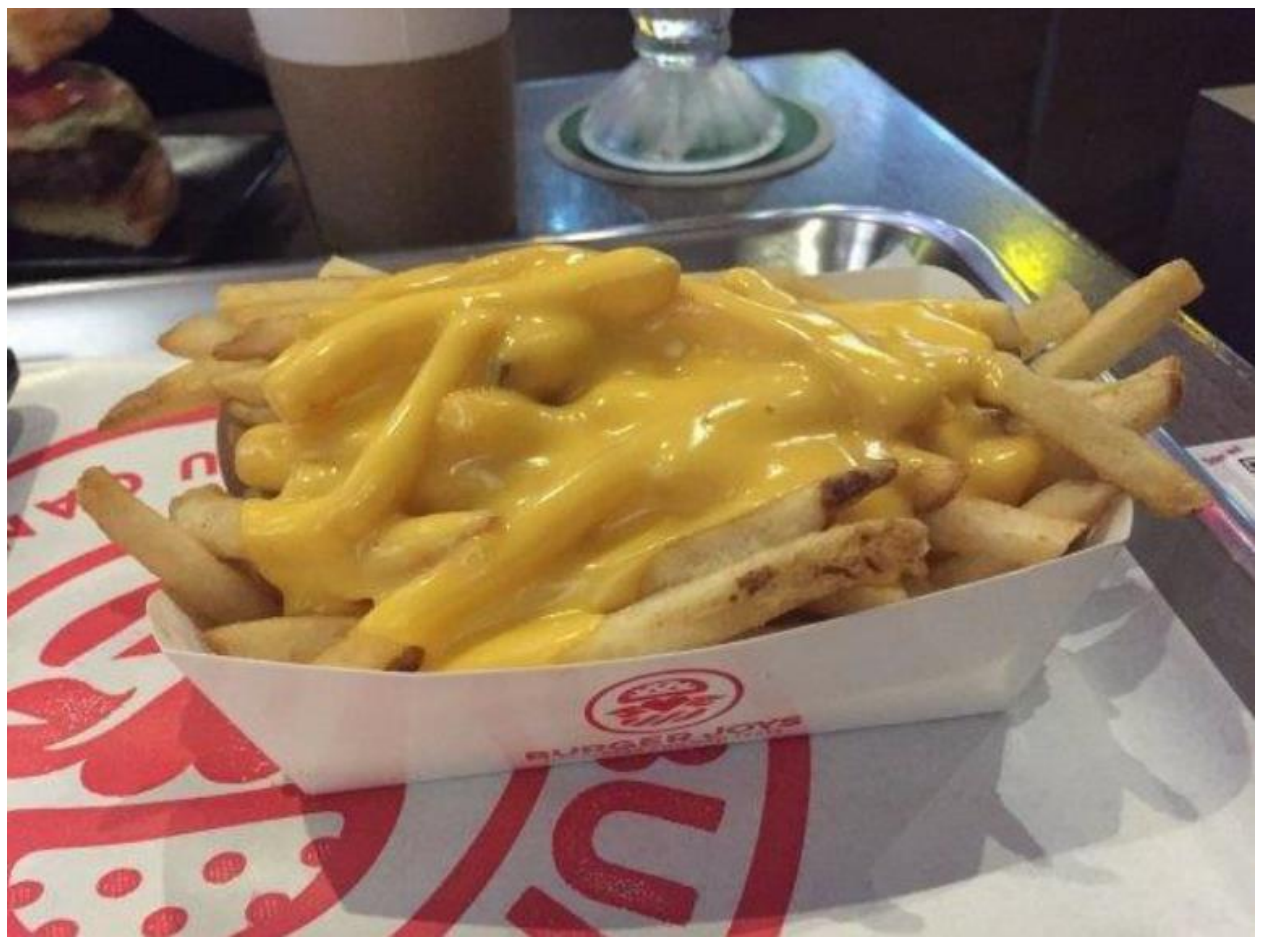

Ambiguous image: ambiguity between creamy_paste or grated

Figure 8: Ambiguous images with high probability of being mislabeled.

In deep learning, higher is the number of data, higher is the accuracy of trained model [24]. The model can be improved by using a larger dataset with enhanced variety of instances.

Other experiments can be done on the model by freezing less number of layers of base model and fine-tuning more layers. In this way, the base model can be trained on the new dataset which can improve overall performance of the model. Further experiments can be done using the combination of SGD with Adam optimizers. The resultant model is not overfitted, so, we can try increasing the learning rate to 0.001 in place of current learning rate 0.0001, for faster convergence while training on higher number of epochs.